\documentclass{llncs}
\usepackage{makeidx}  % allows for indexgeneration
\usepackage{graphicx}
\usepackage{subfigure}
\usepackage{amssymb,amsmath,bm}
\usepackage{mathrsfs}
\usepackage{booktabs}
\usepackage{multirow}
\usepackage{float}
\usepackage{textcomp}
\usepackage{color}
\usepackage{url}
\begin{document}
	\title{FetusMap: Fetal Pose Estimation in 3D Ultrasound}
	
	\author{Xin Yang$^{1}$ \and Wenlong Shi$^{2,3}$ \and Haoran Dou$^{2,3}$ \and Jikuan Qian$^{2,3}$ \and Yi Wang$^{2,3}$ \and Wufeng Xue$^{2,3}$ \and Shengli Li$^4$ \and Dong Ni$^{2,3*}$\and Pheng-Ann Heng$^{1}$}
	
	%index{Xin, Yang}
	%index{Wenlong, Shi}
	%index{Haoran, Dou}
	%index{Jikuan, Qian}
	%index{Yi, Wang}
	%index{Wufeng, Xue}
	%index{Shengli, Li}
	%index{Dong, Ni}	
	%index{Pheng-Ann, Heng}

	\institute{Department of Computer Science and Engineering, The Chinese University of \\Hong Kong, Hong Kong, China\and
		National-Regional Key Technology Engineering Laboratory for Medical Ultrasound, Guangdong Key Laboratory for Biomedical Measurements and Ultrasound Imaging, School of Biomedical Engineering, Health Science Center, Shenzhen University, Shenzhen, China\and
		Medical UltraSound Image Computing (MUSIC) Lab\and
		Department of Ultrasound, Affliated Shenzhen Maternal and Child Healthcare,	Hospital of Nanfang Medical University, Shenzhen, China}
	
	\maketitle
	\let\thefootnote\relax\footnotetext{*Corresponding author: nidong@szu.edu.cn}

	\begin{abstract}
		The 3D ultrasound (US) entrance inspires a multitude of automated prenatal examinations. However, studies about the structuralized description of the whole fetus in 3D US are still rare. In this paper, we propose to estimate the 3D pose of fetus in US volumes to facilitate its quantitative analyses in global and local scales. Given the great challenges in 3D US, including the high volume dimension, poor image quality, symmetric ambiguity in anatomical structures and large variations of fetal pose, our contribution is three-fold. \textbf{(\textit{i})} This is the first work about 3D pose estimation of fetus in the literature. We aim to extract the skeleton of whole fetus and assign different segments/joints with correct torso/limb labels. \textbf{(\textit{ii})} We propose a self-supervised learning (SSL) framework to finetune the deep network to form visually plausible pose predictions. Specifically, we leverage the landmark-based registration to effectively encode case-adaptive anatomical priors and generate evolving label proxy for supervision. \textbf{(\textit{iii})} To enable our 3D network perceive better contextual cues with higher resolution input under limited computing resource, we further adopt the gradient check-pointing (GCP) strategy to save GPU memory and improve the prediction. Extensively validated on a large 3D US dataset, our method tackles varying fetal poses and achieves promising results. 3D pose estimation of fetus has potentials in serving as a map to provide navigation for many advanced studies. \par
	\end{abstract}

	\section{Introduction}
	Featured as real-time and radiation-free, ultrasound (US) is widely accepted in clinic for fetal health monitoring. Plenty of diagnostic biometrics can be automatically interpreted from the US images by recent researches. With broad field-of-view and low user dependency, the advent of 3D US further brings opportunities for automated solutions to attain precise descriptions of fetus \cite{yang2019towards}. \par

	\begin{figure}[h]
		\centering
		\subfigure[ ]{\includegraphics[width=0.30\textwidth]{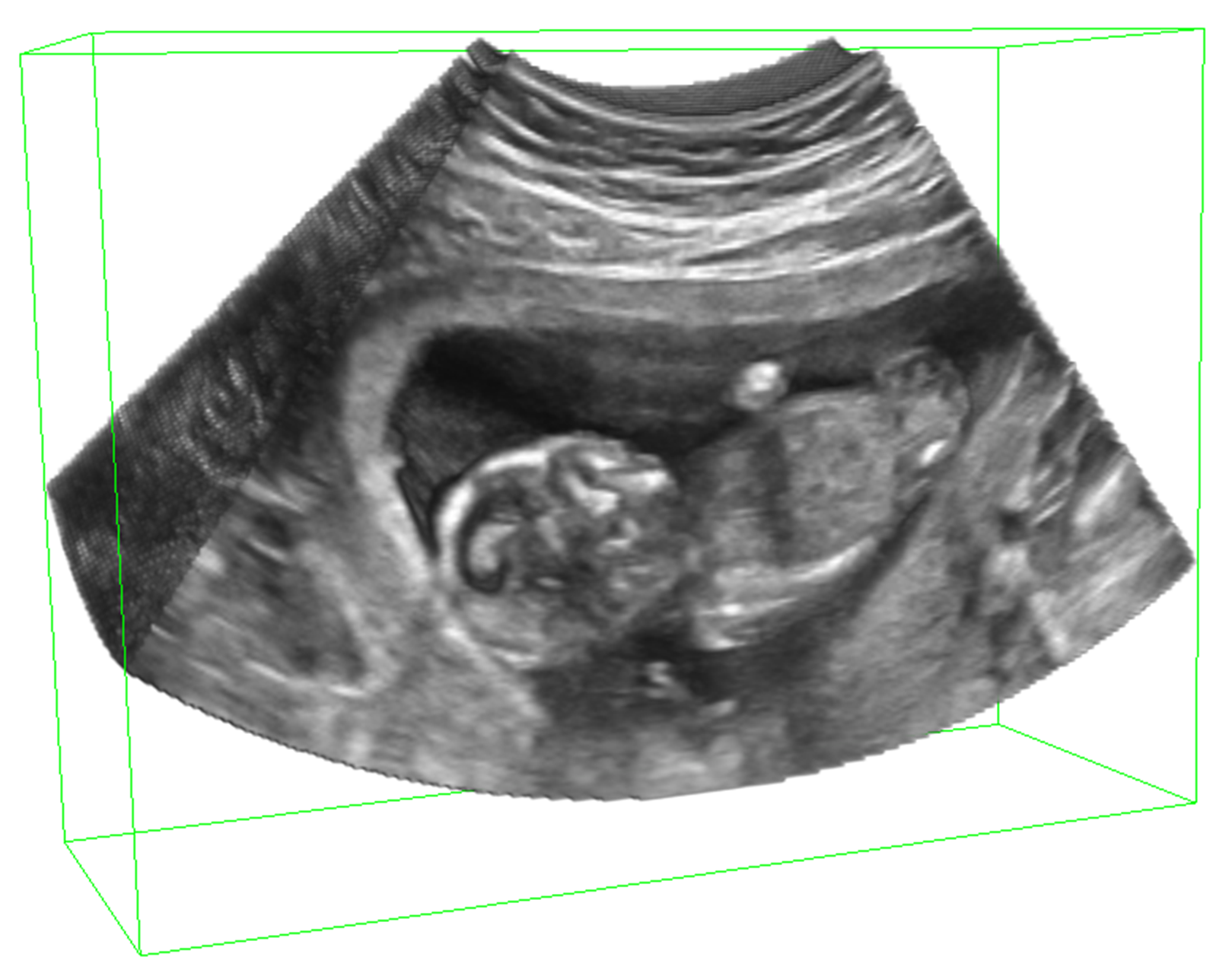}}
		\hspace{.4in}
		\subfigure[ ]{\includegraphics[width=0.18\textwidth]{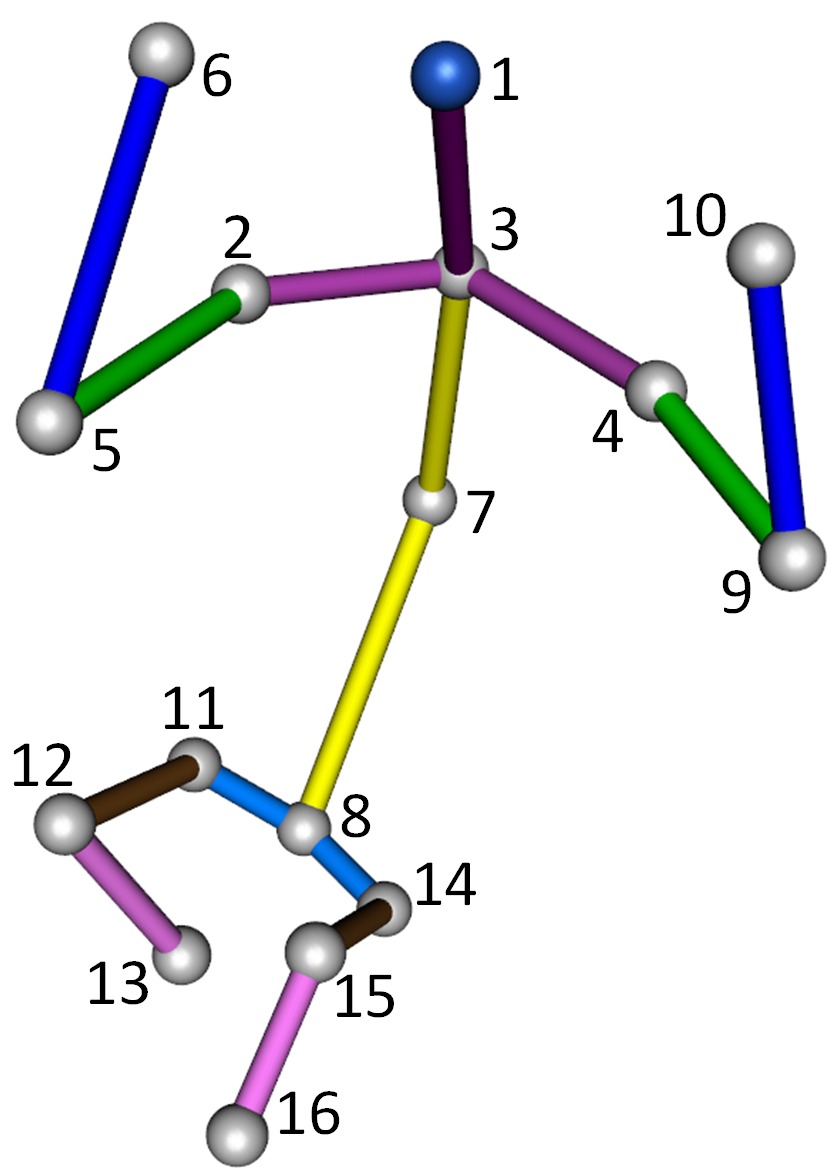}}
		\hspace{.4in}
		\subfigure[ ]{\includegraphics[width=0.20\textwidth]{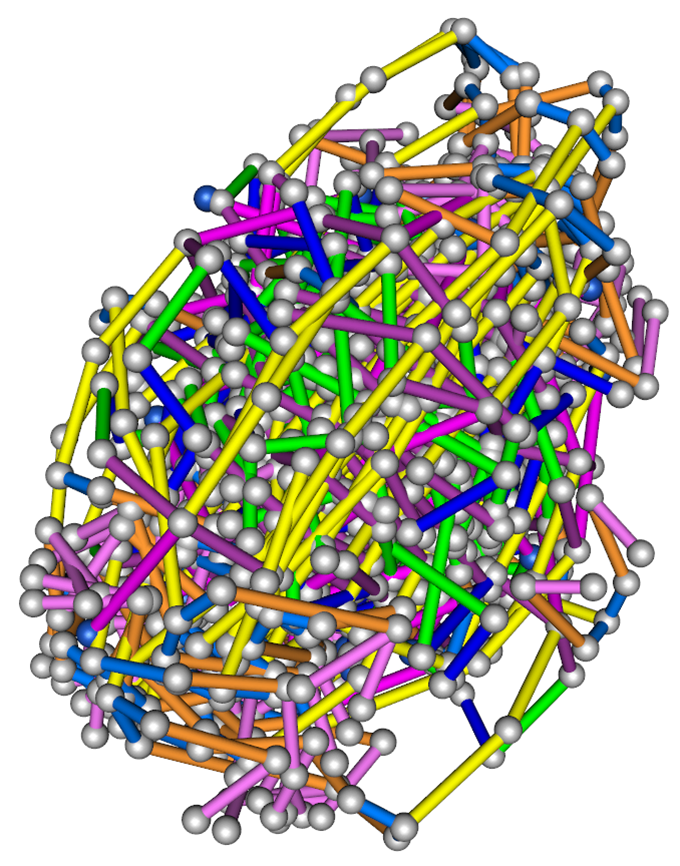}}		
		\caption{3D pose estimation of fetus in US volumes. (a) A sectional view of a fetus in US volume. (b) An instance of 3D fetal pose with 16 landmark indexes and 15 colored segments. (c) All the pose annotations of 152 fetuses in our dataset. Large variations exist when referring to (b). Better view in color version.}
		\label{fig:challenge_show}
		\vspace{-0.5cm}
	\end{figure}
	
	However, currently, there still lacks a solution to provide structuralized descriptions for the whole fetus in 3D US. This description should facilitate not only the traditional tasks in local scale, like standard plane detection \cite{baumgartner2017sononet} and biometric measurements \cite{wu2017cascaded}, but also the advanced analyses in global scale, like fetal movement pattern and longitudinal comparison. Therefore, we propose to approach this goal by exploring a new task dedicated to 3D pose estimation of fetus in US volumes. Specifically, as illustrated in Fig. \ref{fig:challenge_show}(b), by localizing 16 landmarks of fetus in fully body, we aim to extract the skeleton of whole fetus and assign different segments/joints with correct torso/limb labels. \par	

	% Challenges
	As shown in Fig. \ref{fig:challenge_show}, estimating fetal pose in 3D US needs to tackle several challenges. \textit{First}, the image quality of 3D US is relatively low due to the speckle noise, low resolution and acoustic shadows (Fig. \ref{fig:challenge_show}(a)). \textit{Second}, large variations in fetal pose, scale and orientation cause high image appearance variations, which not only generate the ambiguity in localizing symmetric landmarks but also degrade the generalization ability of automated methods. (Fig. \ref{fig:challenge_show}(b)\&(c)). \textit{Third}, accurate landmark localization heavily depends on the global context in the whole volume to suppress false positives. However, digesting the whole US volume with size about 200$\times$200$\times$200 is very tough under limited computing resources. \par
	
	% related work
	Deep neural network is nowadays the dominant method for landmark detection in 3D US. A multi-task deep network was proposed in \cite{namburete2018fully} for fetal eye localization in US volumes.. Huang et al. exploited a semi-supervised learning method to localize 6 fetal head landmarks \cite{huang2018omni}. To use the geometric or class constraints, generative adversarial scheme was explored in \cite{xu2018less} to regularize the landmark predictions. Although these methods are promising, the networks often suffer from their limited generalization ability, especially in our task, where fetuses have free poses with varying appearances. For 2D pose estimation, it has been well studied in computer vision field. Chen et al. proposed to learn the joint inter-connectivity prior in an adversarial scheme to refine the human pose prediction \cite{chen2017adversarial}. Liu et al. further distilled the articulated relationship between joints with recurrent neural network for feature boosting \cite{liu2019feature}. However, facing with the large volume and varying poses, these methods tend to be degraded as our experiments show. \par

	In this paper, we try to tackle the challenges in 3D US for whole-body fetal pose estimation and generalize the landmark detection for large volumes. Our contribution is three-fold. \textbf{(\textit{i})} To the best of our knowledge, this is the first work about 3D pose estimation of fetus in the literature. We believe that taking the fetal pose estimation as a map, navigation can be generated to assist a series of advanced studies on automated prenatal examinations. \textbf{(\textit{ii})} We propose a self-supervised learning (SSL) framework to force the deep network to produce visually plausible pose predictions. Specifically, we leverage the landmark-based registration to effectively encode case-adaptive anatomical priors and generate evolving label proxy for supervision. The proxy is a suboptimal supervision but proves to be explicit in conveying prior knowledge for successive refinement. \textbf{(\textit{iii})} To enable our 3D deep network generate better features with higher resolution input under limited computing resource, we further adopt the gradient check-pointing (GCP) strategy to save GPU memory. With little computation overhead, GCP facilitates the training and inference of larger volumes and hence contributes to better localization performance. With extensive experiments on a large 3D US dataset, our proposed method deals with varying fetal poses and presents to be general with promising results. \par
	
	\begin{figure}[h]
		\centering
		\vspace{-0.6cm}
		\includegraphics[width=0.95\linewidth]{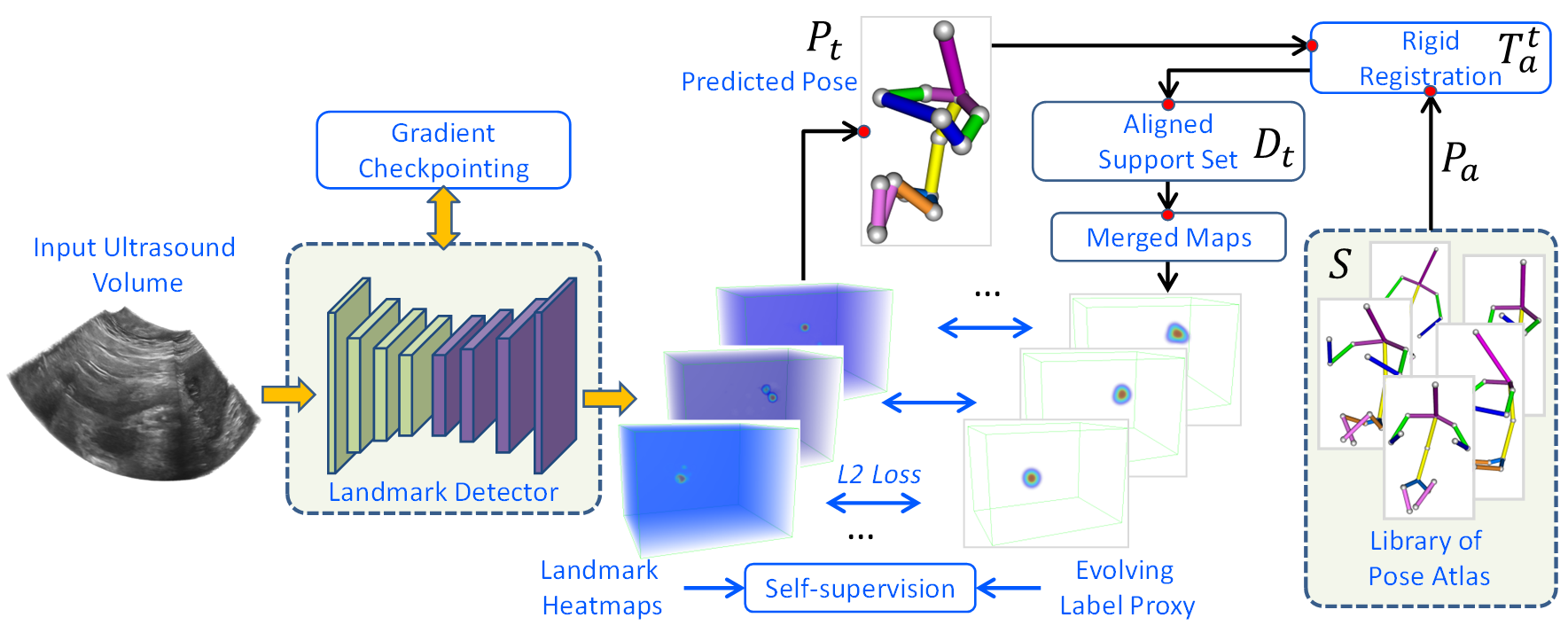}
		\caption{Schematic view of our proposed framework for on-line refinement.}
		\label{fig:framework}
		\vspace{-0.9cm}
	\end{figure}
			
	\section{Methodology}
	Fig. \ref{fig:framework} is the overview of our proposed framework. System input is a whole US volume. A pre-trained deep network based landmark detector firstly digests the input and predicts the heatmap of 16 landmarks with an intermediate fetal pose estimation. By retrieving a support set of atlases in the pose library via rigid registration, label proxies are produced to form the self-supervision. The landmark detector is then tuned iteratively for on-line refinement. The system outputs the final pose estimation after a few number of iterations. Landmark detector is updated under the gradient checkpointing strategy in necessary. \par
	
	\begin{figure}[h]
		\centering
		\includegraphics[width=0.83\textwidth]{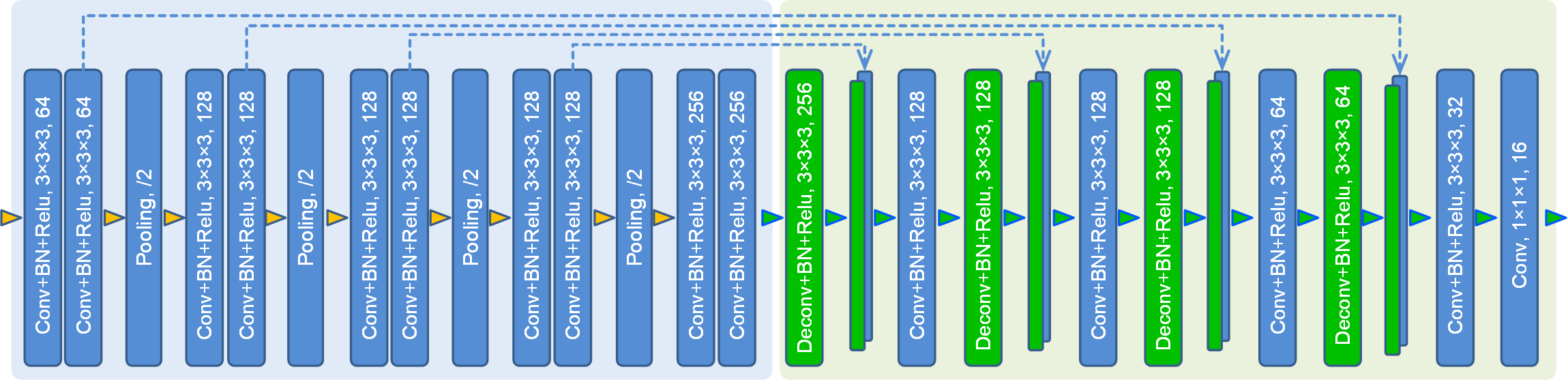}
		\caption{Our proposed U-net like architecture for landmark detection.}
		\label{fig:land_networks}
		\vspace{-0.4cm}
	\end{figure}

	\subsection{Backbone of Landmark Detector}
	Since localizing fetal landmarks needs to consider both the global context and local details, as shown in Fig. \ref{fig:land_networks}, we build a 3D U-net \cite{ronneberger2015u} like network to simultaneously localize 16 landmarks of fetus in full body. Specifically, we deepen the network with consecutive convolutional (Conv) layers in a block and 4 pooling layers to encode high-level semantic features of the whole volume. Each Conv and deconvolutional (Deconv) is followed by a batch normalization (BN) layer and a rectified linear unit (ReLU). L2 regression loss is minimized as loss function. \par
	
	\subsection{Self-supervised Learning for On-line Refinement}
	Due to the large variations of fetal pose, scale and orientation, deep networks for 3D fetal pose estimation in US often suffer from the low generalization ability when facing with varying and unseen fetal appearances. Anatomical prior is helpful for the problem \cite{liu2019feature,chen2017adversarial}. However, these priors are often modeled in an indirect way and hard to take effect in our task (see Section \ref{section:experiment}). In this paper, as shown in Fig. \ref{fig:framework}, we propose to address the problem by producing direct shape prior for on-line refinement under a SSL scheme. \par
	
	Supervising a model with the label proxy generated by the model itself and thus being annotation-free is the core idea of SSL. SSL changes classic testing fashion from simple inference to on-line learning. It fine-tunes the trained deep model with a label proxy. Strong guidance from the label proxy helps deep model update itself and generalize well to unseen cases. Recently, conditional random field \cite{bai2017semi} and interactive annotation \cite{wang2018interactive} have been proposed to learn pixel-wise dependency to synthesize the label proxy in SSL. Whereas, these methods are intractable for our discrete landmark detection. Therefore, we propose to synthesize the landmark label proxy by combining the model prediction with the shape knowledge of a pose library.
	{\setlength\abovedisplayskip{1pt plus 3pt minus 2pt}
	\setlength\belowdisplayskip{1pt plus 3pt minus 7pt}
		\begin{align}
		\label{eq:evolve_1}
		&\mathcal{D}_{t} = \mathop{\arg\min}_{\mathcal{D}^{\prime}\subset\mathcal{S},|\mathcal{D}^{\prime}|=K} \sum_{a=1}^{N}\sum_{j=1}^{16}\parallel \mathrm{T}_{a}^{t}\times\vec{P}_{a}^{j}-\vec{P}_{t}^{j}\parallel_{2},
		\end{align}
	}	
	
	As shown in Fig. \ref{fig:framework}, after being pre-trained on the training dataset, the landmark detector enters our SSL scheme for testing. Following the Eq. \ref{eq:evolve_1}, for an unseen testing US volume of fetus, detector predicts its 16-channel landmark heatmaps and an intermediate 3D pose estimation $\vec{P}_{t}$. Each atlas $\vec{P}_{a}$ in the pose library $\mathcal{S}$ is then aligned to $\vec{P}_{t}$ via a rigid transformation $\mathrm{T}_{a}^{t}$. Since there often exist flaws in the landmarks of pose $\vec{P}_{t}$, we only select a subset of landmarks $\vec{P}_{t}^{j}$ to calculate the $\mathrm{T}_{a}^{t}$. Specifically, referring to Fig. \ref{fig:challenge_show}, we choose the landmarks $j\in\{1,2,3,4,5,7,8,9,11,14\}$ which can be robustly detected across the dataset and also have relatively small variances to fulfil the rigid registration conditions. By retrieving the top-\textit{K} candidates with lowest registration errors, a support set of aligned atlas $\mathcal{D}_{t}$ is formed. $\textit{K}=10$ in this paper. Then, a 16-channel landmark label proxy is produced by averaging the landmark gaussian maps of the aligned atlases in $\mathcal{D}_{t}$. The label proxy will serve as the pseudo ground truth in iteration $t$ to trigger the loss function. Landmark detector needs to update itself to refine its predictions and also the label proxy to minimize the loss. \par
	
	Although the label proxy is initially rough, it encodes case-adaptive and strong shape prior which helps the detector to generalize to unseen US cases. The label proxy will evolve towards a suboptimal and case-specific state as the SSL iterates. Effectiveness of SSL will be elaborated in Section \ref{section:experiment}. \par
	
	\subsection{Enable Better Performance with Larger Input}\label{section:train_test}
	Limited by GPU memory, 3D deep models often sacrifice the input size to enlarge network capacity. Plenty of content details are destroyed during the downscaling. Reducing the GPU memory consumption to break the bottleneck of input size is crucial for our task. In this work, we opt for the gradient checkpointing (GCP) strategy \cite{chen2016training,openai2018} to trade off the GPU memory usage with re-computation and make US volume with high resolution available for deep model. \par
	
	\begin{figure}[h]
		\centering
		\vspace{-0.6cm}		
		\includegraphics[width=0.82\textwidth]{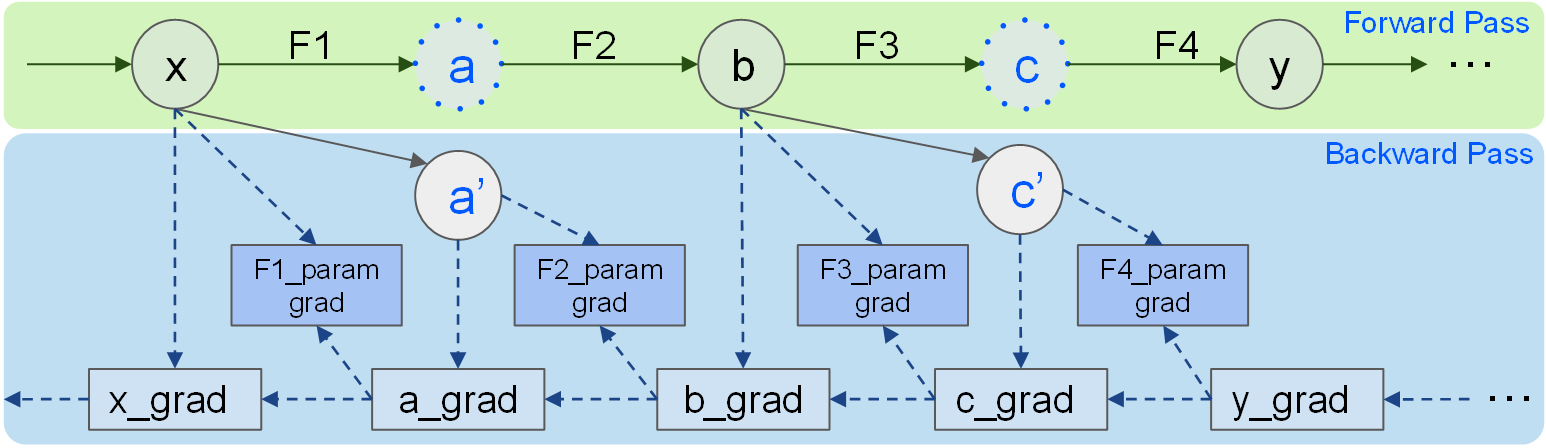}
		\caption{Illustration of the forward pass and gradient re-computation in backward pass of the GCP. Dotted circle denotes the node in the computation graph to be emptied.}
		\label{fig:checkpoint_grad}
		\vspace{-0.4cm}
	\end{figure}	
	
	As shown in Fig. \ref{fig:checkpoint_grad}, the core idea of GCP is discarding the data in some computation graph nodes after approaching a milestone node to make more GPU memory available for subsequent inference. The data of the discarded nodes will then be recovered by re-computation during backpropagation. Given an input $x$, data in node $a$ is computed by the parameters of the $F_{1}$ function. Based on $a$ and $F2$, node $b$ is then approached. At this moment, the data in $a$ will be discarded to release the occupied GPU memory. During the backward pass, to get the gradients of $F_{1}$ and $F_{2}$, node $a$ will be recovered as $a^{\prime}$ by the re-computation from $x$ and $F_{1}$. The gradient for the parameters of $F_{2}$ is calculated from $a^{\prime}$ and the gradient of $b$ ($b\_grad$). With $a^{\prime}$ as a transition node, the gradient for the parameters of $F_{1}$ ($F1\_param  grad$) can be further obtained. Thus, without losing model accuracy, both the forward and backward passes can fit in the GPU. For our task, restricted by the skip connections, we manually select all the Conv layers except the Conv layers directly connected with the concatenation layer into the layer set for GCP. By using GCP to reduce GPU memory consumption, we can enable the network process US volumes with high resolution (enlarged as 1.25 times on each dimension). \par
	
	\section{Experimental Results} \label{section:experiment}
	\subsubsection{Materials and Implementation}
	We validate our method on a dataset of 152 fetal US volumes acquired from 152 pregnant volunteers with gestational age ranges from 10$\sim$14 weeks. Average size of volume is 220$\times$205$\times$260. Voxel size is 0.5$\times$0.5$\times$0.5 mm. Approved by local IRB, all volumes were anonymized and obtained by experts using a Mindray DC-8 system. Free fetal poses are allowed. An expert with 10-year experience manually annotated 16 landmarks. These 16 landmarks cover the fetal head, neck, shoulder, elbow, wrist, spine, sacra, hip joint, knee and ankle. We randomly split the dataset into 100/52 volumes for training/testing. Training set is augmented to 800 with flipping and rotation. \par
	
	We implement our method in \textit{Tensorflow}, using a standard PC with only one NVIDIA TITAN Xp GPU (12GB). \textit{Codes will be online available.} The original US volume is downscaled as \textit{0.4} times before input into our basic landmark detector. \textit{0.4} is the highest ratio allowed by the GPU for our network. With the GCP, we can enlarge the ratio to \textit{0.5}. During the training of landmark detector on the training dataset, we update the weights with an Adam optimizer (batch size=1, initial learning rate is \textit{1e-3}, moment term is 0.5, epoch=20). During the testing with SSL, initial learning rate is decreased to \textit{5e-4}. Landmark detector runs on each testing case with SSL for 6 iterations (about 12 seconds in total). GCP is used for all the methods compared in this paper when it is needed. Training with GCP needs about 1.5 times of extra running time. \par
	
	\subsubsection{Quantitative and Qualitative Analysis}
	Two metrics are used to evaluate accuracy of pose estimation: the Euclidean distance (mm) between landmark prediction and ground truth, and the area under PCK curve (AUC, \%), where PCK is the Percentage of Correct Key points, i.e., the percentage of detections with Euclidean distance below a threshold. With the basic landmark detector (Land) as backbone, we compared our SSL method with two typical refinement methods that explore the landmark dependency: (a) generative adversarial learning (GAN) \cite{chen2017adversarial,xu2018less} and (b) recurrent neural network (RNN) \cite{liu2019feature}. We implemented GAN by learning to classify the pair of US volume and 16-channel heatmaps, and RNN by adding a convolutional RNN layer to the last Conv layer of our landmark detector. GCP is applied to input when the downscale ratio is \textit{0.5}. 
	\begin{table}[!htb] \caption {Comparison of Euclidean Distance in Landmark Localization} \label{table:Euclidean_metric}
		\centering
		\vspace{-0.1cm}
		\scriptsize
		%\tiny
		\begin{tabular}{c|c|c|c|c|c|c|c|c|c|c|c|c|c|c|c|c|c}
			\toprule[2pt]		
			\multirow{2}{*}{\bf{Method}} & \multicolumn{17}{c}{\bf{Euclidean Distance [mm] \textdownarrow}}\\
			\cline{2-18}
			&L1 &L2 &L3 &L4 &L5 &L6 &L7 &L8 &L9 &L10 &L11 &L12 &L13 &L14 &L15 &L16 &\textit{mean}\\
			\hline
			Land-R4    &1.75 &6.69 &2.54 &7.18 & 8.85 & 11.1 & 2.57 & 3.31 & 9.87 & 13.7 & 4.34 & 9.23 & 7.22 & 3.84 & 6.69 & 6.46 &\textit{6.59}\\
			LandGCP    &1.74 &8.78 &2.54 &5.59 &\textcolor{blue}{6.39} & 6.40 &\textcolor{blue}{2.15} & 2.39 & 11.5 & 11.1 & 2.97 &\textcolor{blue}{6.27} & 5.05 & 2.36 & 5.91 & 4.49 &\textit{5.35}\\
			\hline        	
			RNN-R4       &1.85 &12.1 &7.62 &14.6 & 22.1 & 22.6 & 2.70 & 10.4 & 20.2 & 18.9 & 10.7 & 12.3 & 6.86 & 2.87 & 5.49 & 7.11 &\textit{11.14}\\
			RNNGCP       &1.76 &9.47 &5.93 &13.5 & 18.1 & 14.9 & 4.45 & 10.4 & 17.1 & 15.0 & 4.41 & 7.27 & 5.56 & 5.33 & 7.11 & 5.35 &\textit{9.11}\\
			\hline        	
			GAN-R4       &1.85 &7.16 &\textcolor{blue}{2.18} &6.57 & 8.54 & 11.7 & 2.44 & 2.50 & 10.3 & 11.4 & 3.35 & 8.30 &\textcolor{blue}{5.00} & 3.37 & 5.35 & 4.21 &\textit{5.89}\\
			GANGCP       &1.68 &8.61 &2.42 &5.18 & 7.79 & 9.48 & 2.34 & 2.32 & 11.2 & 12.2 & 2.99 & 6.84 & 6.29 & 2.02 & 5.43 &\textcolor{blue}{4.18} &\textit{5.69}\\
			\hline        	
			SSL-R4       &\textcolor{blue}{1.72} &\textcolor{blue}{5.00} &2.36 &\textcolor{blue}{4.37} & 6.81 & 13.3 & 2.56 & 3.19 &\textcolor{blue}{8.40} & 10.8 & 3.32 & 6.45 & 7.40 & 2.93 &\textcolor{blue}{4.63} & 4.26 &\textit{5.47}\\
			SSLGCP       &1.76 &6.39 &2.44 &4.57 &6.66 &\textcolor{blue}{6.00} & 2.23 &\textcolor{blue}{2.30} & 9.27 &\textcolor{blue}{9.40} &\textcolor{blue}{2.65} & 6.51 & 5.68 &\textcolor{blue}{1.98} & 5.60 & 5.31 &\textcolor{blue}{\textit{4.92}}\\
			\toprule[2pt]
		\end{tabular}
		\vspace{-0.3cm}
	\end{table}
	
	\begin{table}[!htb] \caption {Comparison of AUC in Landmark Localization} \label{table:PCK_metric}
		\centering
		\vspace{-0.1cm}		
		\scriptsize
		%\tiny
		\begin{tabular}{c|c|c|c|c|c|c|c|c|c|c|c|c|c|c|c|c|c}
			\toprule[2pt]		
			\multirow{2}{*}{\bf{Method}} & \multicolumn{17}{c}{\bf{AUC Ratio [\%] \textuparrow}}\\
			\cline{2-18}
			&L1 &L2 &L3 &L4 &L5 &L6 &L7 &L8 &L9 &L10 &L11 &L12 &L13 &L14 &L15 &L16 &\textit{mean}\\
			\hline
			Land-R4    &81.4 &56.5 &75.5 &50.1 &47.5 &32.3 &72.8 &74.0 &48.0 &28.4 &63.6 &27.3 &44.9 &61.8 &48.1 &49.4 &\textit{53.8}\\
			LandGCP    &82.5 &48.4 &74.8 &61.0 &\textcolor{blue}{65.6} &51.8 &78.4 &75.9 &45.7 &35.8 &70.4 &46.5 &49.7 &76.6 &51.7 &55.7 &\textit{60.7}\\
			\hline			
			RNN-R4     &80.6 &34.2 &77.1 &29.6 &34.3 &21.9 &71.3 &72.7 &35.5 &29.1 &55.5 &37.3 &\textcolor{blue}{54.7} &69.7 &59.3 &55.6 &\textit{51.2}\\
			RNNGCP     &82.8 &40.3 &\textcolor{blue}{77.3} &36.5 &35.5 &36.2 &75.5 &76.0 &26.6 &27.5 &58.1 &43.9 &51.3 &62.4 &54.0 &\textcolor{blue}{60.0} &\textit{52.8}\\
			\hline			
			GAN-R4     &80.6 &55.5 &77.2 &55.1 &50.3 &33.0 &74.2 &74.4 &46.6 &38.1 &66.6 &32.9 &51.9 &66.4 &55.0 &56.4 &\textit{57.1}\\
			GANGCP     &\textcolor{blue}{83.4} &49.8 &75.6 &64.2 &57.7 &36.1 &76.4 &\textcolor{blue}{77.0} &46.6 &35.8 &71.1 &44.6 &51.6 &80.0 &55.8 &58.8 &\textit{60.3}\\
			\hline			
			SSL-R4     &81.8 &\textcolor{blue}{61.9} &75.9 &65.6 &56.7 &22.0 &73.0 &75.3 &\textcolor{blue}{56.8} &37.8 &65.1 &46.8 &42.4 &70.1 &\textcolor{blue}{63.0} &57.0 &\textit{59.5}\\
			SSLGCP     &82.6 &57.7 &75.6 &\textcolor{blue}{66.1} &63.4 &\textcolor{blue}{55.0} &\textcolor{blue}{77.5} &76.9 &56.0 &\textcolor{blue}{43.6} &\textcolor{blue}{73.6} &\textcolor{blue}{47.4} &45.1 &\textcolor{blue}{80.5} &55.3 &50.5 &\textcolor{blue}{\textit{62.9}}\\
			\toprule[2pt]
		\end{tabular}
		\vspace{-0.2cm}
	\end{table}
	
	\begin{figure}[h]
		\centering
		\includegraphics[width=0.94\linewidth]{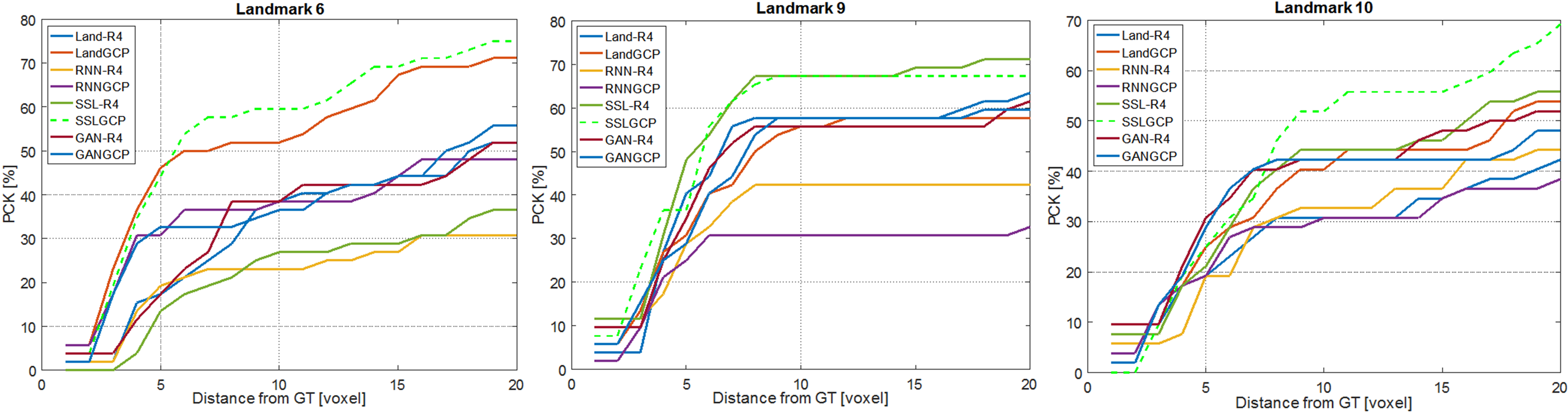}
		\caption{PCK curves for 3 fetal landmarks. x axis is the distance threshold. \textit{SSLGCP} (dotted green curve) gets the best results among all the competitors.}
		\label{fig:PCK_curves}
	\end{figure}
	
	Table \ref{table:Euclidean_metric} presents the Euclidean distance of different methods for all the 16 landmarks. We use \textit{R4} to denote the model handling input with downscale ratio of \textit{0.4}, and \textit{GCP} the method with GCP to handle input with larger downscale ratio of \textit{0.5}. As demonstrated in the table,  almost all methods achieved lower prediction distance for all the landmarks with GCP, benefiting from its better features perceiving from higher resolution input. With this work, we are the first to prove that, GCP can improve landmark localization by enabling larger ultrasound volume input. Besides, although RNN and GAN based refinement methods bring improvements over the $Land$, they still perform obviously worse for some landmarks. With the case-adaptive label proxy as a strong prior, SSL based methods surpass the GAN/RNN and get almost the best results by achieving the top rank on 10 landmarks. The advantage of SSL can also be drawn from the average prediction distance, according to which the proposed SSL achieves an average distance of 4.92mm, and significantly outperforms the two competitors.\par
	
	\begin{figure}[h]
		\centering
		\subfigure[ ]{\includegraphics[width=0.482\textwidth]{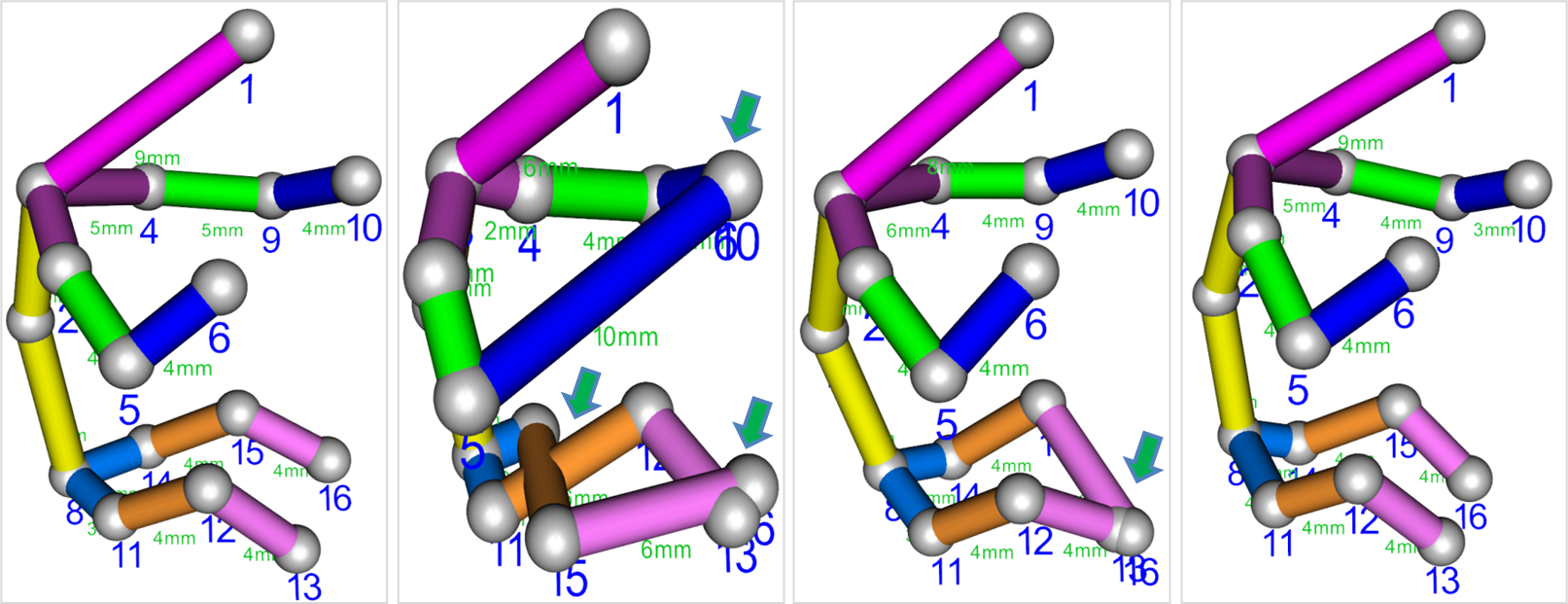}}
		\hspace{.1in}
		\subfigure[ ]{\includegraphics[width=0.46\textwidth]{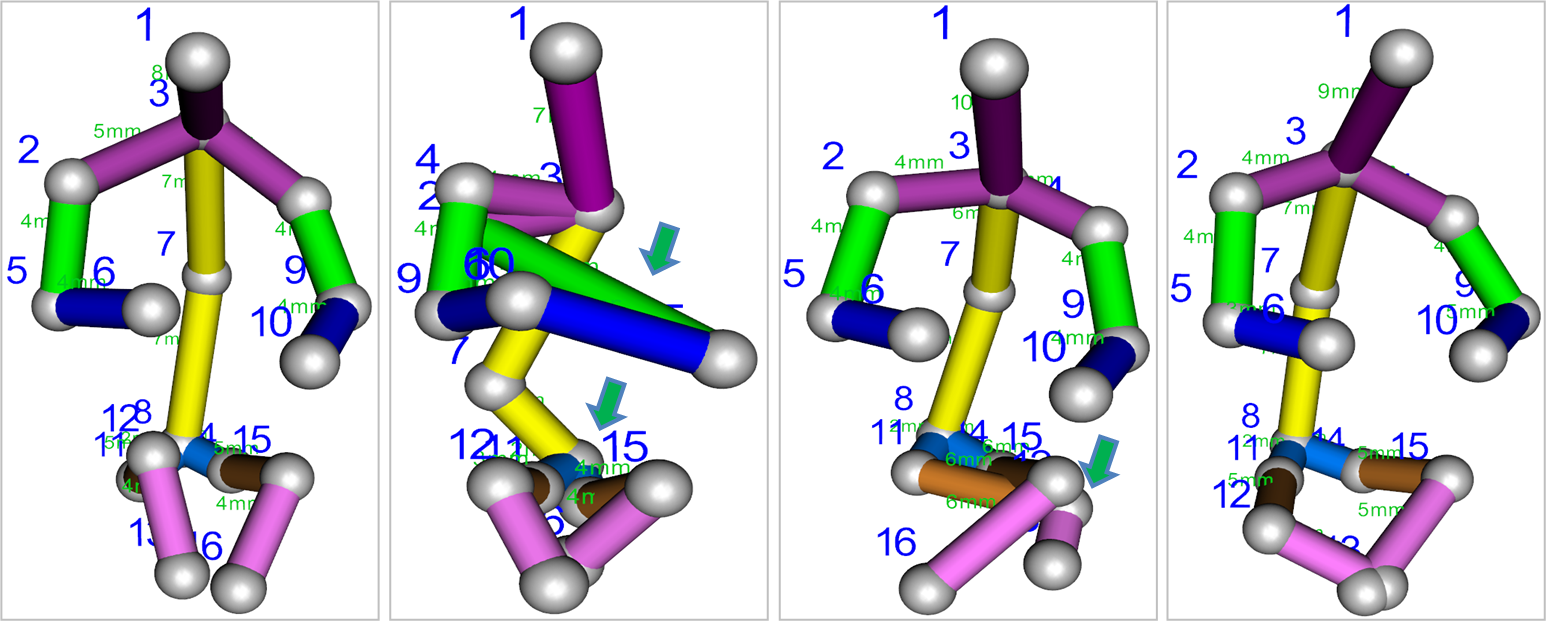}}
		\caption{Visualization of two 3D fetal pose estimations. From left to right: ground truth, \textit{Land-R4}, \textit{LandGCP} and \textit{SSLGCP}. Blue digit for landmark index, green digit for length.}
		\label{fig:visualization}
		\vspace{-0.5cm}
	\end{figure}

	PCK evaluates the distribution of predicted landmarks around ground truth. Table \ref{table:PCK_metric} further compares the AUC of methods. Similar trends for GCP and SSL can be observed. SSL equipped with GCP (\textit{SSLGCP}) tops the task of most landmark detections. It also achieves the highest mean AUC among all competitors. The highest improvement over the baseline \textit{Land-R4}, about $20\%$, occurs on the detection of landmarks \textit{L4, L6, L10, L12} and $L14$. Referring to Fig. \ref{fig:challenge_show}, we can find that these are the symmetric landmarks on the limb which are hard to be differentiated by \textit{Land, RNN} and \textit{GAN} methods. We believe that both the strong shape prior from the evolving label proxy and the better feature input enabled by the GCP contribute to this significant improvement. We further provide the PCK curves of these landmarks from different methods in Fig. \ref{fig:PCK_curves} for readers to check details. \par
	
	In Fig. \ref{fig:visualization}, we visualize two cases of fetal pose estimations to show the advantages of our method \textit{SSLGCP}. \textit{Land-R4} and \textit{LandGCP} tend to be trapped by symmetric landmarks (green arrows), while our method can rectify these flaws and presents visually plausible estimations. As a byproduct of the pose estimation, the lengths of key segments of fetus are also produced in the 3D pose. \par
	
	\section{Conclusion}
	In this paper, we propose the first work about 3D fetal pose estimation in US volumes. We mainly tackle the challenges from the generalization ability with self-supervised learning and computation burden of large volumes with gradient checkpointing strategy. Extensive experiments prove the feasibility and effectiveness of our proposed method. We believe the pose estimation of fetus can serve as map and inspire the automated prenatal US image analyses. \par
	
	\subsubsection{Acknowledgments:}
	The work in this paper was supported by the grant from Research Grants Council of Hong Kong SAR (Project No. CUHK14225616), National Natural Science Foundation of China(Project No. U1813204) and Shenzhen Peacock Plan (No. KQTD2016053112051497, KQJSCX20180328095606003). \par
	
	% ---- Bibliography ----
	%
	\bibliographystyle{splncs}
	\bibliography{refs}
	
\end{document}